\documentclass[letterpaper]{article} 
\usepackage[]{aaai2026}  
\usepackage{times}  
\usepackage{helvet}  
\usepackage{courier}  
\usepackage[hyphens]{url}  
\usepackage{graphicx} 
\usepackage{natbib}  
\usepackage{caption} 
\usepackage{algorithm}
\usepackage{algorithmic}
\usepackage{tabularx}
\usepackage{xcolor}
\usepackage{amsfonts}
\usepackage{amsmath}
\usepackage{booktabs}
\usepackage{subcaption}
\usepackage{multirow}

\usepackage{newfloat}
\usepackage{listings}
\DeclareCaptionStyle{ruled}{labelfont=normalfont,labelsep=colon,strut=off} 
\floatstyle{ruled}
\newfloat{listing}{tb}{lst}{}
\floatname{listing}{Listing}
\title{Vulnerability-Aware Robust Multimodal Adversarial Training}
\author{
    Junrui Zhang\textsuperscript{\rm 1}, 
    Xinyu Zhao\textsuperscript{\rm 2}, 
    Jie Peng\textsuperscript{\rm 1}, 
    Chenjie Wang\textsuperscript{\rm 3}, \\
    Jianmin Ji\textsuperscript{\rm 1,}\thanks{Corresponding Author}, 
    Tianlong Chen\textsuperscript{\rm 2}
}
\affiliations{
    \textsuperscript{\rm 1}University of Science \& Technology of China,\\
    \textsuperscript{\rm 2}University of North Carolina at Chapel Hill\\
    \textsuperscript{\rm 3}Institute of Artificial Intelligence, Hefei Comprehensive National Science Center \\

    \{zhangjunrui, pengjieb\}@mail.ustc.edu.cn, wangchenjie@iai.ustc.edu.cn,\\
    \{xinyu,tianlong\}@cs.unc.edu, jianmin@ustc.edu.cn
}

\usepackage{bibentry}

\usepackage{xspace}  
\newcommand{\ours}{\texttt{VARMAT}\xspace}

\begin{document}

\maketitle

\begin{abstract}

Multimodal learning has shown significant superiority on various tasks by integrating multiple modalities.
However, the interdependencies among modalities increase the susceptibility of multimodal models to adversarial attacks.
Existing methods mainly focus on attacks on specific modalities or indiscriminately attack all modalities. 
In this paper, we find that these approaches ignore the differences between modalities in their contribution to final robustness, resulting in suboptimal robustness performance.
To bridge this gap, we introduce \textbf{V}ulnerability-\textbf{A}ware \textbf{R}obust \textbf{M}ultimodal \textbf{A}dversarial \textbf{T}raining (\texttt{VARMAT}), a probe-in-training adversarial training method that improves multimodal robustness by identifying the vulnerability of each modality.
To be specific, \texttt{VARMAT} first explicitly quantifies the vulnerability of each modality, grounded in a first-order approximation of the attack objective (Probe). Then, we propose a targeted regularization term that penalizes modalities with high vulnerability, guiding robust learning while maintaining task accuracy (Training).
We demonstrate the enhanced robustness of our method across multiple multimodal datasets involving diverse modalities.
Finally, we achieve $\{12.73\%, 22.21\%, 11.19\%\}$ robustness improvement on three multimodal datasets, revealing a significant blind spot in multimodal adversarial training.

\end{abstract}

\begin{links}
    \link{Code}{https://github.com/AlniyatRui/VARMAT}
\end{links}


\section{Introduction}

With the rapid advancement of artificial intelligence, multimodal learning~\cite{lianghigh, yucrema, lei2024vit, cafead} has demonstrated powerful perception and reasoning capabilities across a wide range of real-world applications.
These abilities are particularly valuable in complex and dynamic environments, where relying on a single modality often fails to provide sufficient information for robust and reliable prediction.
For instance, in sentiment analysis, the same sentence can express different emotions depending on how it is spoken, where multimodal features are necessary to provide a comprehensive understanding of the speaker's intent~\cite{zadeh2018multimodal}.

Nevertheless, the advantages of multimodal learning are accompanied by significant security challenges. It is widely recognized that deep neural networks are vulnerable to adversarial examples crafted by adding human-imperceptible perturbations, which mislead models into making incorrect predictions~\cite{GoodfellowSS14, madry2018towards}. While this vulnerability is inherent in most deep learning models, recent studies indicate that incorporating multiple heterogeneous modalities substantially increases the attack surface~\cite{bagdasaryan2024adversarial, qi2024visual}, providing adversaries more opportunities for exploitation. For example, adversarial illusions~\cite{bagdasaryan2024adversarial} have been proposed to construct perturbations that make a target modality's embedding close to an arbitrary, adversary-chosen input in another modality, resulting in conflicts between modalities during task execution. 

\begin{figure}[t]
    \centering
    \includegraphics[width=\linewidth]{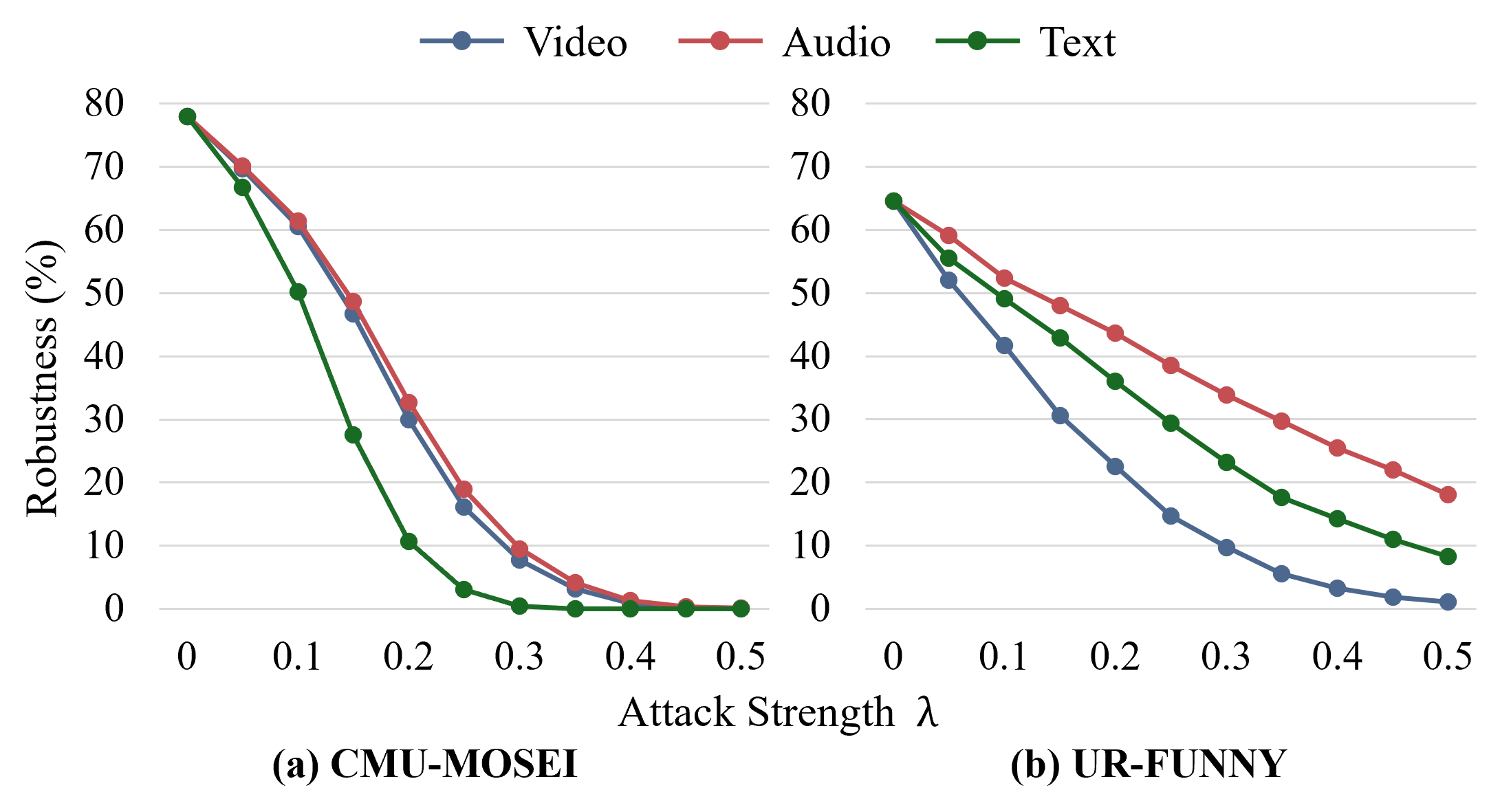}
    \caption{Adversarial robustness under varying attack strengths $\lambda \in [0,0.5]$ for different modalities, showing significant differences in modality-specific vulnerabilities.}
    \label{fig:attack_robustness}
\end{figure}

Therefore, it is necessary to design a multimodal defense strategy rather than a single-modal defense to protect against potential adversarial attack threats from arbitrary modalities. 
Researchers have explored methods to improve the robustness of multimodal models~\cite{luo2023image,li2024one,zhangclipure}, among which adversarial training has emerged as one of the most popular defense strategies due to its effectiveness and practicality~\cite{zhang2019theoretically, pangbag}.
Although existing methods have achieved promising results through elaborate attack strategies and adversarial training frameworks, we observe that most of them attack each modality indiscriminately, neglecting the inherent heterogeneity of different modalities~\cite{yin2023vlattack, gao2024boosting, zheng2024unified, zhangrethinking}. 
Moreover, the design of input-level defenses typically relies on well-defined constraints and standardized input formats, which are often unavailable for some emerging modalities, like the tabular modality in the MIMIC dataset~\cite{johnson2016mimic}, or for modalities that use pre-extracted features, such as point clouds or other 3D structures commonly used to accelerate training~\cite{yucrema}.

To address these issues, we explore multimodal robustness through feature-space perturbation. This approach offers a unified view, allowing us to focus on the robustness of each modality without being concerned about the heterogeneities between them.
Specifically, given the latent feature of all modalities, we apply consistent constraints and attack algorithms to these features to evaluate the vulnerabilities of each modality individually.
As shown in Figure~\ref{fig:attack_robustness}, we conduct single-modality PGD attacks~\cite{madry2018towards} on the HighMMT model~\cite{lianghigh}. We use the CMU-MOSEI~\cite{zadeh2018multimodal} and UR-FUNNY~\cite{urfunny} datasets, perturbing one modality at a time to assess the model's robustness.
The results show that the model exhibits varying levels of robustness when facing identical adversarial attacks across different modalities in the feature space, indicating that some modalities are more vulnerable than others.
This suggests that adversarial training strategies should consider the heterogeneities between modalities, particularly those modalities that are more susceptible to attacks, to achieve better robustness.

Therefore, we first propose a vulnerability-aware attack method to demonstrate that leveraging this discrepancy can craft stronger adversarial examples. 
Our method reallocates perturbation budgets based on estimated vulnerability, measured by feature magnitude and gradient norm,
grounded in a first-order approximation of the attack objective.
The perturbation budget refers to the constraint on the maximum allowable perturbation applied to the feature space. By concentrating perturbations on more susceptible modalities, the attack achieves stronger performance, while also revealing the importance of modality-specific vulnerabilities for improving adversarial robustness. 
Building on this insight, we further propose \ours, a vulnerability-aware robust multimodal adversarial training method that directly optimizes the gradient norms of all modalities to dynamically balance and reduce their vulnerabilities, thereby enhancing robustness against adversarial attacks.  
To be specific, we do not use feature magnitude during training, as directly regularizing it may compromise the expressive capacity of each modality and lead to overfitting under certain attack constraints.

Finally, we evaluate our method on three multimodal datasets from the MultiBench Benchmark~\cite{liang2021multibench} using the HighMMT~\cite{lianghigh} under various adversarial training settings.
In particular, we adopt fast adversarial training methods throughout our experiments, due to their favorable trade-off between robustness and computational efficiency~\cite{shafahi2019adversarial, wongfast, wang2024preventing, jia2022prior}.
The results demonstrate the effectiveness and efficiency of \ours. Specifically, \ours achieves robustness improvements of $\{12.73\%, 22.21\%, 11.19\%\}$ on the three datasets, respectively, revealing a critical blind spot in current multimodal adversarial training and underscoring the importance of vulnerability-aware, differentiated optimization to improve robustness.
The primary contributions of our research are outlined as follows:
\begin{itemize}
    \item We systematically reveal the inherent discrepancies in adversarial robustness across different modalities and approximate the attack objective to identify the critical factors that contribute to each modality’s vulnerability.

    \item Building on the identified critical factors, we design a vulnerability-aware attack strategy that reallocates the perturbation budget to concentrate on more vulnerable modalities, achieving better attack performance and underscoring the importance of modality-specific vulnerabilities for improving adversarial robustness.

    \item  We propose \ours, a novel vulnerability-aware robust multimodal training method that addresses the previous neglect of inherent discrepancies across modalities. \ours incorporates a targeted regularization term that adaptively balances and suppresses the vulnerability of each modality, resulting in better robustness while maintaining the model's expressive capacity.
    
    \item Extensive experiments on multiple multimodal datasets demonstrate the effectiveness and efficiency of \ours, revealing a significant blind spot in multimodal adversarial training.
    
\end{itemize}

\begin{figure*}[t]
    \centering
    \includegraphics[width=\linewidth]{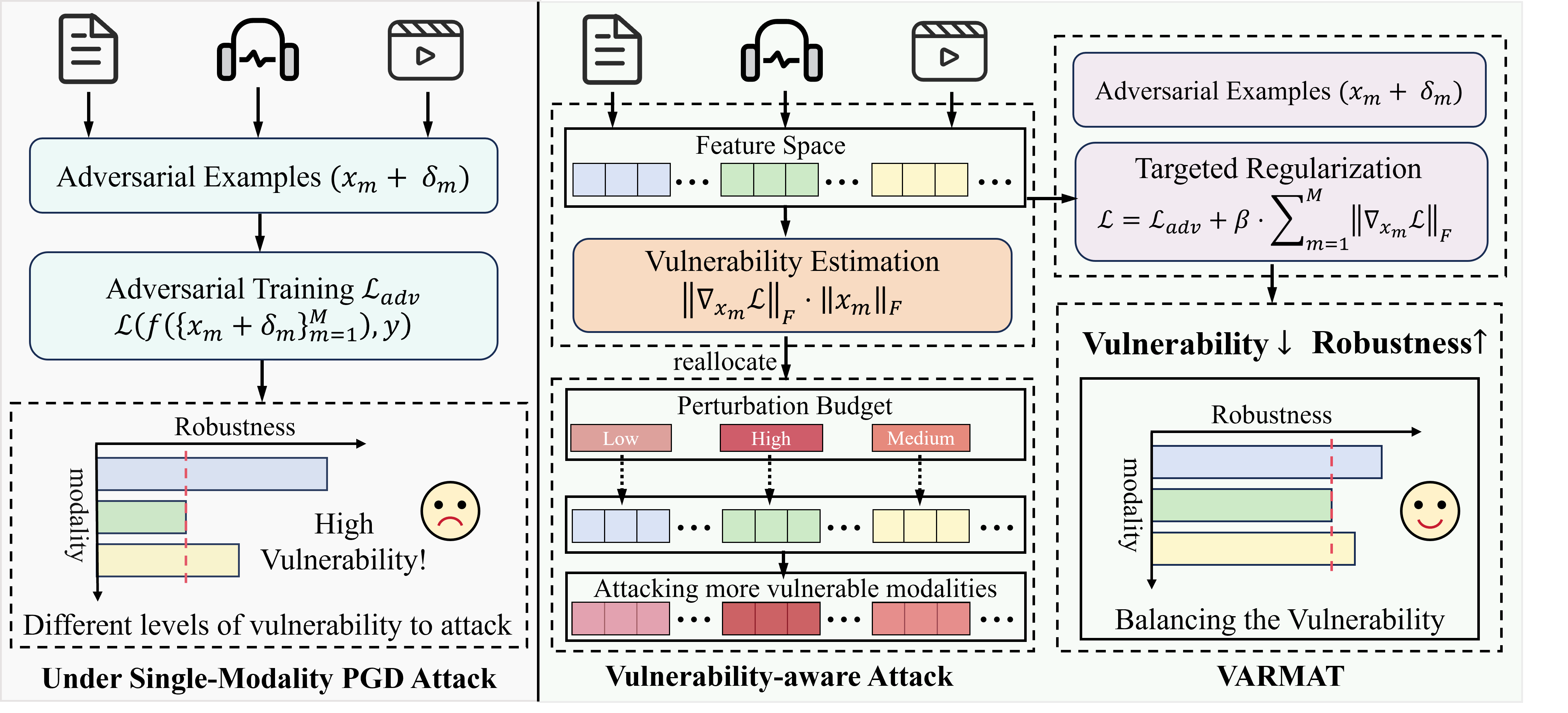}
    \caption{Comparison of vulnerability-aware attack and \ours with previous indiscriminate adversarial training.}
    \label{fig:overview}
\end{figure*}

\section{Related Work}
\subsection{Multimodal Adversarial Attack}

Adversarial attacks refer to deliberately crafted inputs designed to mislead machine learning models into making incorrect predictions. Among various attack strategies, gradient-based methods have emerged as some of the most straightforward and effective approaches, primarily due to their direct exploitation of model vulnerabilities via gradients.
FGSM~\cite{GoodfellowSS14} pioneered this line of research by demonstrating that even a single-step perturbation along the gradient direction could significantly degrade model performance.
Building upon this, iterative variants such as I-FGSM~\cite{kurakin2018adversarial} were developed to generate more potent adversarial examples through fine-grained, multi-step optimization. To address the challenge of poor local optima, MI-FGSM~\cite{dong2018boosting} introduced momentum to stabilize gradient updates across iterations, thereby improving attack transferability. Further improvements were achieved by GI-FGSM~\cite{wang2024boosting}, which incorporates a pre-attack phase to estimate a more informed initial perturbation direction, underscoring the critical role of initialization in crafting effective adversarial examples. 

In the context of multimodal learning, multimodal attack strategies exploit interactions between modalities to craft stronger adversarial examples. Co-Attack~\cite{zhang2022towards} establishes a framework for synchronized vision-language perturbations through gradient alignment across modalities, demonstrating superior attack success rates compared to unimodal baselines. VLAttack~\cite{yin2023vlattack} advances this paradigm by introducing a blockwise similarity attack strategy to generate image perturbations that disrupt universal representations. SGA~\cite{lu2023set} incorporates data augmentation during iterative attack generation, significantly improving cross-model transferability through diversified gradient estimation. VLPTransferAttack~\cite{gao2025boosting} further enhances transferability by analyzing the intersections of adversarial trajectories between clean and perturbed inputs, thereby preventing overfitting to a specific threat model. The most recent approach, Anyattack~\cite{zhang2025anyattack}, overcomes the scalability limitations of previous methods by allowing any image to serve as a target for attack without requiring label supervision, achieving better performance.
While these methods demonstrate effectiveness in vision-language settings, several fundamental limitations persist. First, existing works predominantly focus on image-text pairs, neglecting emerging multimodal systems incorporating audio, video, 
and other modalities. Second, they largely overlook the strategic question of where to apply perturbations for maximum impact. Current strategies often attack a single predefined modality or indiscriminately perturb all modalities, implicitly assuming that each modality contributes equally to the model's decision and its adversarial vulnerability. In this work, we reveal the inherent differences in vulnerability across modalities, highlighting their differential impact on adversarial robustness.

\subsection{Fast Adversarial Training}

The most popular defense strategy is PGD-based Adversarial Training~\cite{madry2018towards}, due to its effectiveness and practicality, where adversarial examples generated using multi-step attacks are used to train a robust model. However, the prohibitive cost of standard PGD-based Adversarial Training motivated a search for more efficient alternatives. This led to the development of Fast Adversarial Training~\cite{wongfast}, a class of methods aiming to achieve robust models at a computational cost comparable to standard training using FGSM as attack methods.
FGSM-RS~\cite{wongfast} attempts to mitigate catastrophic overfitting by using a simple initialization with random perturbations.
FGSM-GA~\cite{andriushchenko2020understanding} further improves fast adversarial training by enhancing gradient alignment between clean samples and adversarial examples.
FGSM-PCI~\cite{jia2022prior} proposes that prior knowledge of adversarial perturbations can effectively guide subsequent adversarial training by using the previous epoch's perturbation as the initialization for further epochs, achieving stable and effective optimization directions.
The most recent in this line of research, FGSM-PCO~\cite{wang2024preventing}, introduces a tailored loss function within the training framework to facilitate the recovery of an overfitted model for effective training, thereby preventing the collapse of the inner optimization loop. 
While Fast Adversarial Training has matured significantly in the context of unimodal models, particularly in computer vision, the principles and techniques are now being considered for the next frontier of AI: multimodal models.

\section{Methodology}

In this section, we first introduce the setup of our work in Section~\ref{subsec:preliminary}. Then we identify the critical factors contributing to the differences in vulnerability across modalities in Section~\ref{subsec:l2_constraint}. Futhermore, we propose a vulnerability-aware attack method to highlight the impact of varying vulnerabilities on robustness and present \ours to achieve a stronger defense in Section~\ref{subsec:attack_defense}.

\begin{figure}[t]
    \centering
    \includegraphics[width=\linewidth]{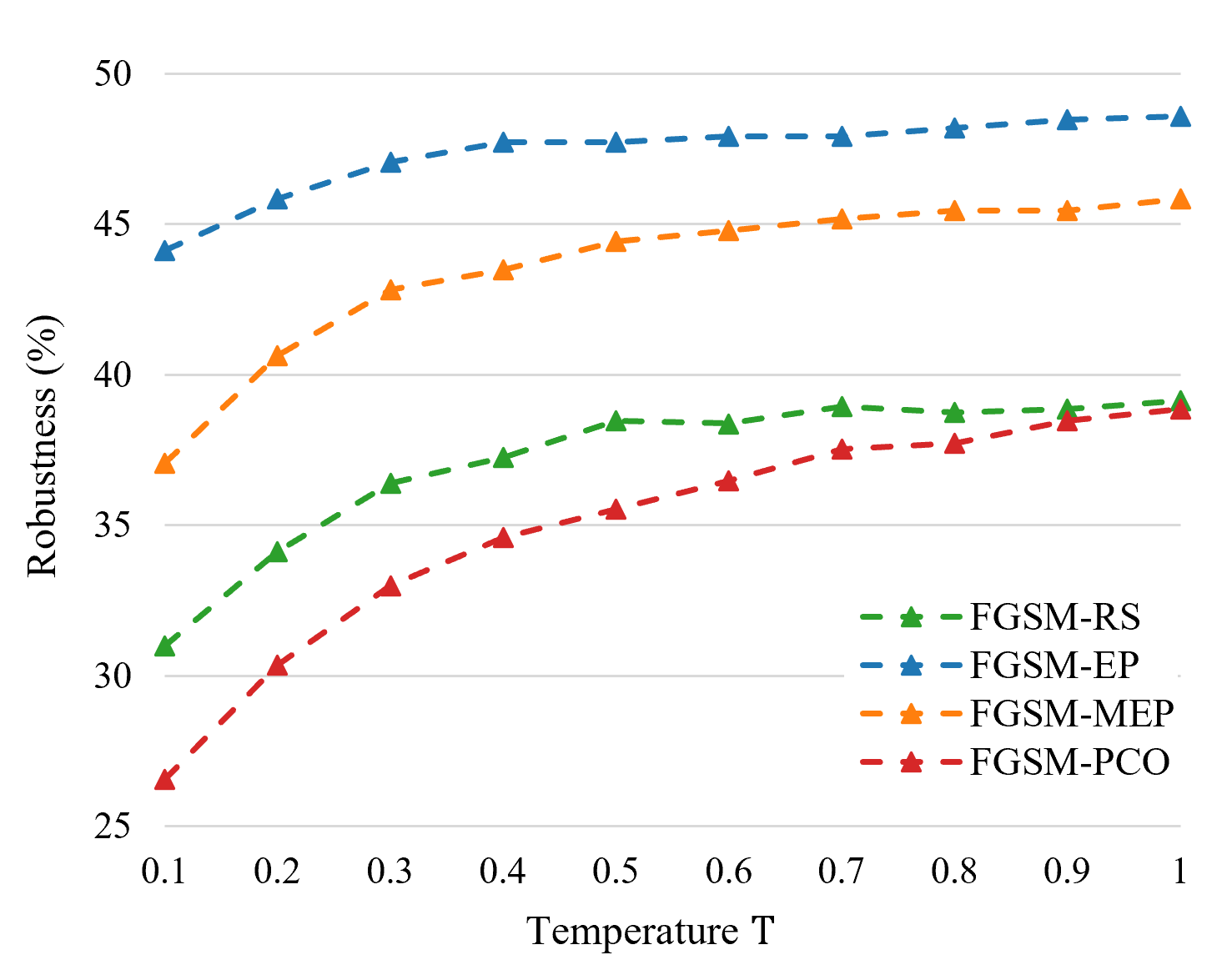}
    \caption{Comparison of robustness between different methods in different temperatures.}
    \label{fig:temperature}
\end{figure}

\subsection{Preliminary}
\label{subsec:preliminary}
Formally, let $f_\theta(\mathrm{x}) \to {y}$ denote a multimodal model, where $\theta$ represents the model parameters, $\mathrm{x} = \{x_1, x_2, \dots, x_M\}$  comprises inputs from $M$ modalities, and $y$ is the ground-truth label. 
Adversarial training is typically formulated as a Min-Max optimization problem:
\begin{equation}
    \arg \min_\theta \; \mathbb{E}_{(\mathrm{x}, y)} \; \arg \max_{\|\delta\|_p \leq \epsilon} \; \mathcal{L}(f_\theta(\mathrm{x} + \delta), y),
\end{equation}

where the inner maximization aims to generate imperceptible perturbations $\delta = \{\delta_1, \delta_2, \dots, \delta_M\}$  that mislead the model's prediction, and the outer minimization leverages these adversarial examples as training samples to improve the model's robustness against adversarial attacks. The constraint $\|\delta\|_p \leq \epsilon$ would project $\mathrm{x} + \delta$ to the $\ell_p$ ball around $\mathrm{x}$ with radius $\epsilon$.
In this paper, we construct perturbations in the feature space from a unified perspective, where the input $\mathrm{x}$ represents encoded features across different modalities.


To thoroughly investigate the varying vulnerabilities across different modalities, we simultaneously study multimodal adversarial attacks and defenses. We start by analyzing the approximation attack objective of inner maximization of multimodal fast adversarial training in Section~\ref{subsec:l2_constraint}. Motivated by this analysis, we reallocate the perturbation budget across modalities based on modality vulnerability weights $w_m$ to construct the vulnerability-aware adversarial attack. The overall constraint becomes:
\begin{equation}
\label{sum_weight}
    \|\delta_m\|_p \leq w_m \cdot \epsilon_m, \quad \text{s.t.}  \sum_{m=1}^M w_m = 1,
\end{equation}

However, in fast adversarial training, recent approaches often leverage prior information
to prevent catastrophic overfitting. Therefore, directly allocating the perturbation budget across epochs may lead to unstable and suboptimal training. To address this issue, we introduce a regularization term based on the theoretical approximation attack rather than directly allocating the perturbation budget. 
Detailed attack and defense strategies are described in Section~\ref{subsec:attack_defense}.

\subsection{Approximation for $Frobenius$-norm Attacks}
\label{subsec:l2_constraint}

Following prior work~\cite{zhufreelb}, we constrain multimodal feature-space perturbations by the $Frobenius$-norm:
\begin{equation}
\epsilon_m = \lambda \cdot \| {x_m} \|_F,
\end{equation}
where $\lambda$ represents the strength of the attack. We then consider a first-order approximation of the inner maximization objective in Eq. (1).
\begin{equation}
    \mathcal{L}(f_\theta(\mathrm{x} + \delta), y) \approx \mathcal{L}(f_\theta(\mathrm{x}), y) + \delta \cdot \nabla_{\mathrm{x}}\mathcal{L}(f_\theta(\mathrm{x}), y),
    \label{eq:adv_loss}
\end{equation}

For simplicity, we denote $\mathcal{L}(f_\theta(\mathrm{x}), y)$ as $\mathcal{L}$. The gradient of a natural sample is used in fast adversarial training to identify the most vulnerable direction for generating single-step attacks. Thus, the first-order term $\delta \cdot \nabla_{\mathrm{x}} \mathcal{L}(f_\theta(\mathrm{x}), y)$ in Eq.~\eqref{eq:adv_loss} approximates the loss increase, denoted as $\Delta \mathcal{L}$.  Since feature are not explicitly constrained by upper or lower bounds, under the $Frobenius$-norm constraint, the single-step perturbation of modality $m$ can be formulated as:
\begin{equation}
\begin{split}
\delta_m &= \frac{\nabla_{x_m}\mathcal{L}}{\|\nabla_{x_m}\mathcal{L}\|_F} \cdot \epsilon_m \cdot w_m,
\end{split}
\end{equation}

Consequently, the approximate loss increase $\Delta \mathcal{L}$ can be expressed as:
\begin{equation}
\label{vulnerablity}
\begin{split}
\Delta \mathcal{L} = \sum_{m=1}^{M}(\|\nabla_{x_m}\mathcal{L}\|_F \cdot \|x_m\|_F \cdot \lambda \cdot w_m) ,
\end{split}
\end{equation}

We observe that modality vulnerability is jointly influenced by both input features and their corresponding gradients. As illustrated in Figure~\ref{fig:overview}, previous methods typically adopt indiscriminate strategies that neglect inherent differences among modalities. To highlight this issue, we design a vulnerability-aware adversarial attack that reallocates the perturbation budget based on modality-specific vulnerabilities, resulting in stronger attacks. Motivated by this insight, we further propose \ours, which employs targeted vulnerability-aware regularization to balance vulnerabilities across modalities during adversarial training, thereby enhancing overall robustness while preserving the model’s expressive capacity.

\subsection{Vulnerability-Aware Adversarial Rosbustness}
\label{subsec:attack_defense}

According to Eq.~\eqref{vulnerablity}, we approximate each modality's vulnerability using the $\|\nabla_{x_m}\mathcal{L}\|_F \cdot \|x_m\|_F$. This estimation can be efficiently computed, as both $x_m$ and  $\nabla_{x_m} \mathcal{L}$ can be obtained through a single forward and backward pass, providing an efficient and lightweight way to probe the relative vulnerability across modalities. Moreover, the probing process is independent of the attack method, which ensures the generality and flexibility of \ours.
Consequently, based on this estimation, we compute the modality-specific vulnerability weights $w_m$ using the following normalization:
\begin{equation}
\begin{split}
w_m= \frac{\|\nabla_{x_m}\mathcal{L}\|_F \cdot \|x_m\|_F}{\sum_{k=1}^M(\|\nabla_{x_k}\mathcal{L}\|_F \cdot \|x_k\|_F)},  
\end{split}
\end{equation}

To further enhance controllability, we apply a temperature-scaled softmax to the normalized weights:
\begin{equation}
w_m = \text{Softmax}\left({w_m}/{T}\right)
\end{equation}

where $T$ is a temperature hyperparameter that modulates the sharpness of the distribution. As $T \to 0$, the perturbation budget becomes increasingly concentrated on the most vulnerable modality; as $T \to \infty$, the weights approach a uniform distribution.
As shown in Figure~\ref{fig:temperature}, we adjust the temperature to examine how adversarially trained models perform under different perturbation allocation patterns. The results indicate that indiscriminate training methods struggle to accommodate modality-specific vulnerabilities across heterogeneous modalities.

Ideally, directly applying this attack strategy during adversarial training could help the model improve its robustness on more vulnerable modalities due to the adaptive reallocation strategy. Although such a strategy was applicable in earlier methods, it conflicts with recent fast adversarial training techniques, which typically incorporate prior knowledge across training epochs to refine the attack process and mitigate catastrophic overfitting.

\begin{algorithm}[tb]
\caption{Vulnerability-Aware FGSM-RS}
\label{alg:algorithm}
\textbf{Input}:Target model $f_\theta$, training epoch $\mathcal{T}$, mini-batch data $\mathcal{B}$, modalties $M$, learning rate $\gamma$, attack strength $\lambda$, regularization coefficient $\beta$.
\\
\textbf{Output}: Model parameter $\theta$
\begin{algorithmic}[1] 
\FOR{$t = 1, \dots, \mathcal{T}$}
    \FOR{$\{\mathrm{x}, y\} \sim \mathcal{B}$}
    \FOR{$m = 1,$ \dots, $M$}
        \STATE $\epsilon_m \leftarrow \frac{\lambda}{M} \cdot \|x_m\|_F$
        \IF{$t == 1$}
            \STATE $\delta_{m} \leftarrow \mathcal{U}_{[-\epsilon_m, \epsilon_m]}$
            \STATE $\delta_m \leftarrow \delta_m \cdot \min\left(1, {\epsilon_m}/{\|\delta_m\|_F}\right)$
        \ENDIF
        \ENDFOR
    \STATE $\mathcal{L}_{Reg} \leftarrow 0$
    \FOR{$m = 1,$ \dots, $M$}
        \STATE $g_t \leftarrow {\nabla_{x_m}\mathcal{L}(f_\theta(\mathrm{x}), y)}$
        \STATE $g_{adv} \leftarrow  {\nabla_{x_m}\mathcal{L}(f_\theta(\mathrm{x}+\delta), y)}$
        \STATE $\delta_m \leftarrow \delta_m + g_{adv}/\|g_{adv}\|_F \cdot \epsilon_m$
        \STATE $\delta_m \leftarrow \delta_m \cdot \min\left(1, {\epsilon_m}/{\|\delta_m\|_F}\right)$
        \STATE $\mathcal{L}_{Reg} \leftarrow \mathcal{L}_{Reg} + \beta \cdot \|g_t\|_F$
    \ENDFOR
    \STATE $\displaystyle \theta_{t+1} \leftarrow \theta_t + \frac{\gamma}{|\mathcal{B}|} \sum^{\mathcal{B}} \nabla \mathcal(\mathcal{L}(f_\theta(\mathrm{x}+\delta), y)+  \mathcal{L}_{Reg}) $
    \ENDFOR
\ENDFOR
\STATE \textbf{return} $\theta$
\end{algorithmic}
\end{algorithm}

To avoid this conflict, we propose incorporating a regularization term that directly optimizes the approximated attack objective $\sum_{m=1}^M(\|\nabla_{x_m}\mathcal{L}\|_F \cdot \|x_m\|_F)$ in Eq.~\eqref{vulnerablity}. Since different modalities contribute unequally to this term, more vulnerable modalities naturally receive stronger optimization weight. As a result, the model implicitly prioritizes these modalities during training, promoting a better alignment between vulnerability and robustness learning.
However, this formulation introduces a potential optimization trap: minimizing the regularization objective may lead to a degenerate solution where $\|x_m\|_F \to 0$. Such collapse impairs the model’s ability to maintain expressive representations and also increases the risk of overfitting. To mitigate this issue, we redesign the regularization term to focus solely on the gradient component, thereby avoiding direct penalization of the feature norm. The revised formulation is:
\begin{equation}
\mathcal{L}_{Reg} = \beta \cdot \sum_{m=1}^M\|\nabla_{x_m}\mathcal{L}\|_F
\end{equation}
where the coefficient $\beta$ controls the strength of regularization. As shown in Algorithm~\ref{alg:algorithm}, we provide a deployment using FGSM-RS, where the fast adversarial training framework is designed to be easily adapted to other algorithms.


\section{Experiments}
\subsection{Experimental Setup}
\subsubsection{Datesets and Model}
Our experiments are conducted on three diverse multimodal datasets from the MultiBench benchmark~\cite{liang2021multibench}: CMU-MOSEI~\cite{zadeh2018multimodal}, UR-FUNNY~\cite{urfunny}, and AVMNIST~\cite{vielzeuf2018centralnet}. CMU-MOSEI and UR-FUNNY include video, text, and audio modalities, while AVMNIST consists of image and audio modalities. For all experiments, we utilize HighMMT~\cite{lianghigh} as our backbone.

\subsubsection{Compared Methods}

We integrate our method with current state-of-the-art fast adversarial training approaches to demonstrate its effectiveness, including FGSM-RS~\cite{wongfast}, FGSM-EP~\cite{jia2022prior}, FGSM-MEP~\cite{jia2022prior}, and FGSM-PCO~\cite{wang2024preventing}. Except for FGSM-RS, other methods incorporate prior information from earlier training epochs.

\subsubsection{Training Details}
For all compared methods, the attack strength is set to $\lambda = 0.01$ during training and $\lambda \in \{0.2, 0.5\}$ during testing.
For multi-step attacks, the step size is set to $\alpha = \epsilon / {I}$, with $I = 10$ iterations. 
We select the best model checkpoints based on the highest robustness on the validation set under $\lambda = 0.2$, evaluated using the PGD attack.
Following the original configurations, we use the Adam optimizer with a learning rate of $0.0008$ and a weight decay of $0.01$. All models are trained on 1 NVIDIA A30 GPU for 10 epochs.
The hyperparameters used in the fast adversarial training methods are consistent with their original settings. The regularization coefficient $\beta$ in our method is determined based on the gradient magnitude of the undefended model trained on each dataset. Specifically, we set $\beta = 1000$ for CMU-MOSEI, and $\beta = 50$ for both UR-FUNNY and AVMNIST.

\begin{table*}[t]
\small
\centering
\begin{tabular}{c | c | c | c c c c | c c c c}
\toprule
\multirow{2}{*}{Method} & \multirow{2}{*}{Variant} & \multirow{2}{*}{Clean} 
& \multicolumn{4}{c|}{CMU-MOSEI ($\lambda=0.2$)} 
& \multicolumn{4}{c}{CMU-MOSEI ($\lambda=0.5$)} \\
\cmidrule(lr){4-7} \cmidrule(lr){8-11}
& & & FGSM & PGD & V-FGSM & V-PGD & FGSM & PGD & V-FGSM & V-PGD \\
\midrule
Undefended       & -         & 77.92 & 27.78 & 22.87   & 26.45 & 21.69 & 0.11  & 0.00& 0.00  & 0.00 \\
\midrule
\multirow{2}{*}{FGSM-RS}    
            & baseline & 78.83 & 72.00 & 71.53  & 71.85 & 71.40 & 57.72 & 55.76& 56.99 & 55.12 \\
            & \ours      & 79.15 & 75.51 & 75.17  & 74.35 & 74.03 & 69.80 & 68.49& 66.57 & 64.91 \\
\midrule
\multirow{2}{*}{FGSM-EP}    
            & baseline & 79.09 & 73.75 & 73.51  & 73.01 & 72.50 & 65.13 & 63.77& 60.78 & 58.86 \\
            & \ours      & 78.07 & 75.17 & 74.86  & 73.81 & 73.40& 70.23 & 68.58 & 66.64 & 65.00 \\
\midrule
\multirow{2}{*}{FGSM-MEP}   
            & baseline & 78.70 & 73.72 & 73.21  & 72.17 & 71.63 & 65.17 & 63.79& 60.41 & 58.65 \\
            & \ours      & 79.30 & 75.64 & 75.36  & 74.50 & 74.05 & 69.63 & 67.46& 66.40 & 64.20 \\
\midrule
\multirow{2}{*}{FGSM-PCO}   
            & baseline & 79.26 & 73.34 & 72.91  & 70.84 & 70.13 & 63.13 & 61.73& 56.26 & 53.63 \\
            & \ours      & 78.72 & 74.28 & 74.03  & 71.20 & 70.64 & 67.28 & 65.58& 59.72 & 58.15 \\
\bottomrule
\end{tabular}
\caption{Robustness (\%) under different attack settings on CMU-MOSEI dataset. Baseline refers to the model trained with standard adversarial training without incorporating \ours.}
\label{tab:fat_comparison_mosei}
\end{table*}

\begin{table*}[t]
\small
\centering
\begin{tabular}{c | c | c c  c  c c | c  c c c c}
\toprule
\multirow{2}{*}{Method} & \multirow{2}{*}{Variant} & \multicolumn{5}{c|}{UR-FUNNY ($\lambda=0.5$)} & \multicolumn{5}{c}{AVMNIST ($\lambda=0.5$)} \\
\cmidrule(lr){3-7} \cmidrule(lr){8-12} 
& & Clean & FGSM & PGD & V-FGSM & V-PGD & Clean & FGSM & PGD & V-FGSM & V-PGD \\
\midrule
Undefended         & -        & 64.56 & 7.09  & 5.20  & 5.48  & 3.78  & 69.04 & 2.57 & 0.12 & 0.74 & 0.00 \\
\midrule
\multirow{2}{*}{FGSM-RS}    & baseline        & 65.69 & 42.25 & 40.45 & 39.79 & 38.19 & 69.38 & 6.11 & 1.78 & 3.23 & 0.33 \\

                            & \ours  & 63.80 & 57.75 & 56.43 & 55.86 & 53.97 & 63.21 & 8.36 & 3.08 & 4.02 & 0.80 \\
\midrule
\multirow{2}{*}{FGSM-EP}    & baseline        & 64.27 & 50.38 & 49.43 & 48.87 & 47.73 & 67.72 & 4.00 & 0.35 & 1.55 & 0.06 \\
                            & \ours  & 62.76 & 57.75 & 57.18 & 56.05 & 54.54 & 62.16 &15.16 &11.08 &12.74 & 8.53 \\
\midrule
\multirow{2}{*}{FGSM-MEP}   & baseline        & 63.80 & 48.20 & 46.50 & 46.60 & 44.52 & 68.98 & 7.37 & 1.22 & 3.65 & 0.22 \\
                            & \ours  & 63.33 & 55.48 & 54.25 & 53.31 & 52.36 & 63.40 &10.18 & 4.50 & 7.79 & 2.63 \\
\midrule
\multirow{2}{*}{FGSM-PCO}   & baseline        & 62.48 & 44.05 & 41.21 & 40.55 & 35.82 & 68.10 & 4.77 & 1.17 & 1.95 & 0.37 \\
                            & \ours  & 63.89 & 59.83 & 59.55 & 58.60 & 58.03 & 66.76 &12.46 & 5.70 & 6.22 & 0.84 \\
\bottomrule
\end{tabular}
\caption{Robustness (\%) under different attack settings on UR-FUNNY and AVMNIST datasets.}
\label{tab:fat_comparison_other}
\end{table*}


\subsubsection{Evaluation Metrics}

We evaluate the robustness of all adversarially trained models under FGSM and PGD attacks.
To be specific, following the standard robust evaluation ~\cite{croce2020reliable}, we compute robustness as the ratio of samples that are correctly classified on both the clean and adversarial examples over the entire dataset, in order to avoid overestimating robustness.


\begin{figure}[t]
    \centering
    \includegraphics[width=\linewidth]{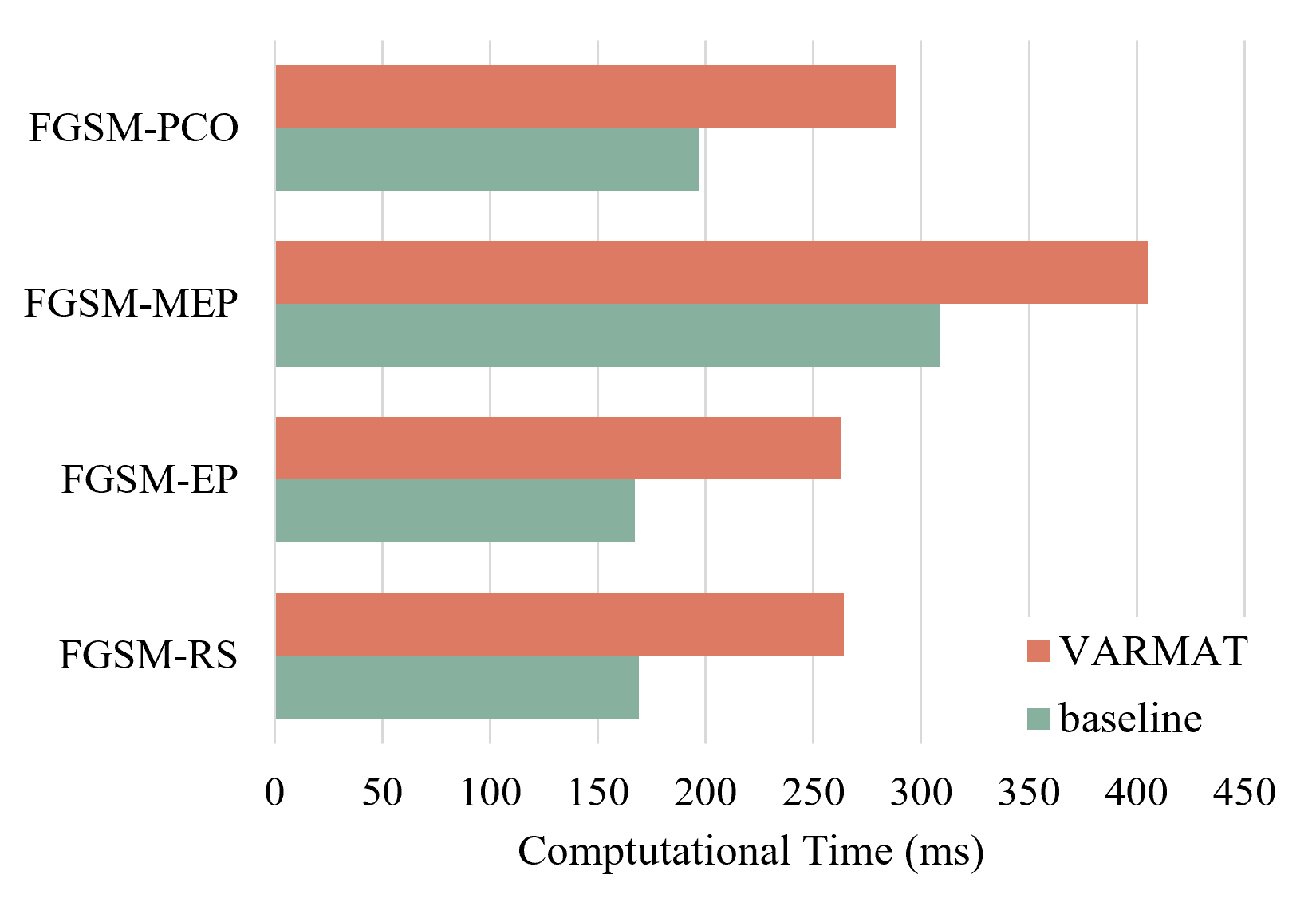}
    \caption{Comparison of computational time between fast adversarial training methods.}
    \label{fig:computational_time}
\end{figure}

\begin{figure}[t]
    \centering
    \includegraphics[width=\linewidth]{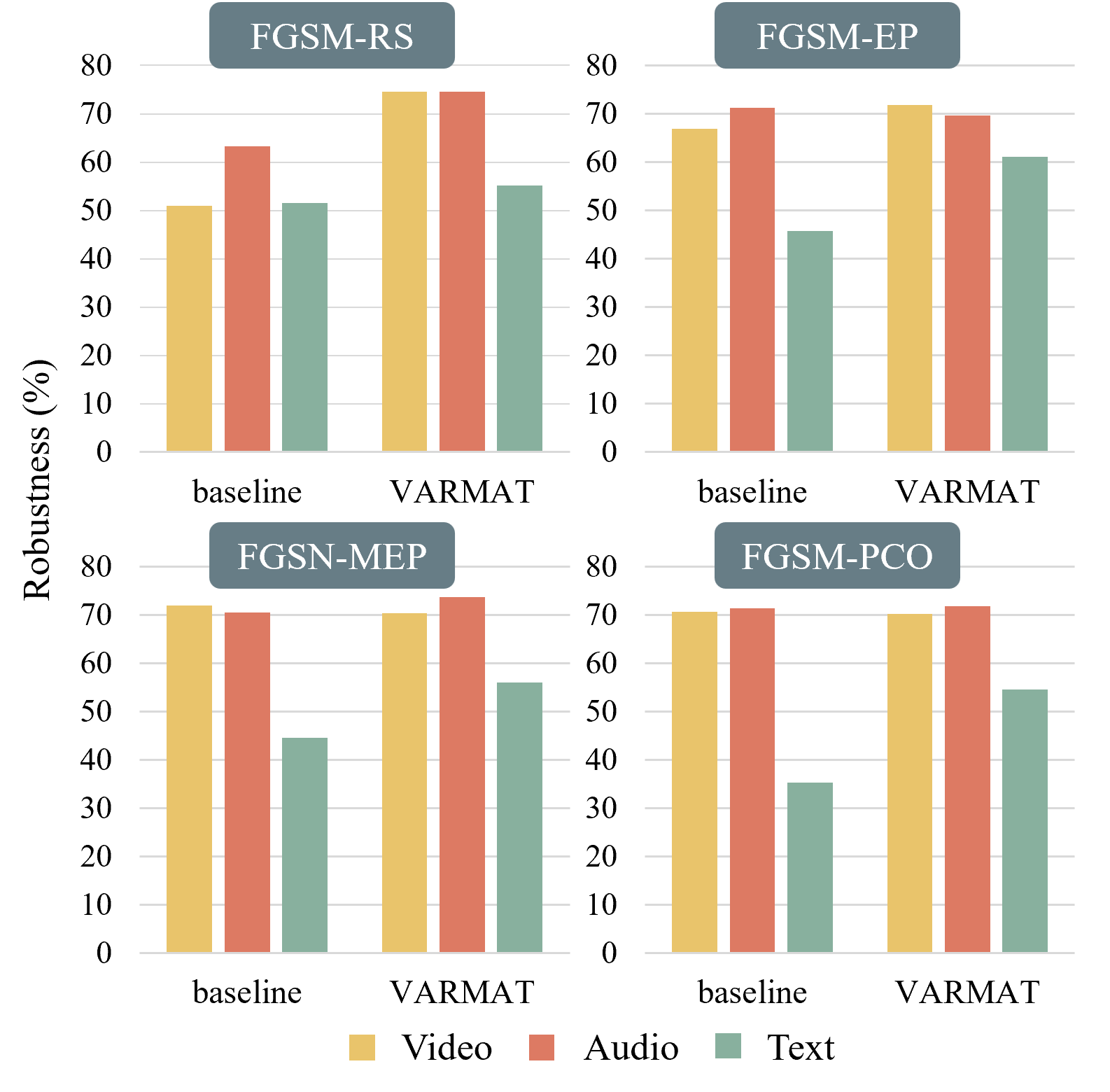}
    \caption{Comparison of single-modality PGD attack performance across fast adversarial training methods ($\lambda=0.5$).}
    \label{fig:balance}
\end{figure}

\subsection{Comparative Results}

\subsubsection{Results on Multimodal Datasets} To demonstrate the effectiveness of \ours, we compare several state-of-the-art fast adversarial training algorithms on the CMU-MOSEI, UR-FUNNY, and AVMNIST datasets.
The comparison results on CMU-MOSEI are shown in Table~\ref{tab:fat_comparison_mosei}, where the baseline refers to the model trained with standard adversarial training without incorporating \ours. We report the robustness under different attack settings as well as the accuracy on the clean samples. Additionally, we evaluate the robustness of each adversarial training method against our proposed vulnerability-aware attack, denoted as V-FGSM and V-PGD, with the temperature $T=0.5$.
Overall, \ours achieves consistently higher robustness across varying attack strengths and attack methods.
The improvements on the four methods reach up to $\{12.73\%, 6.14\%, 5.99\%, 4.52\%\}$, respectively,
which demonstrates that \ours effectively enhances robustness by quantifying and regularizing the vulnerability of each modality.
Moreover, our vulnerability-aware attack leads to a more severe degradation in robustness for both clean and adversarially trained models, highlighting the significance of vulnerability differences and their impact on model robustness.
To further validate the generalizability of our method, we conduct additional experiments on the UR-FUNNY and AVMNIST datasets where \ours achieves consistent robustness improvements on these datasets. As shown in Table~\ref{tab:fat_comparison_other}, in the UR-FUNNY four methods improvements reach up to $\{16.07\%, 7.75\%, 7.84\%, 22.21\%\}$ and in the AVMNIST four methods improvements reach up to $\{2.25\%, 11.19\%, 4.14\%, 7.69\%\}$.

\subsubsection{Single-Modality Attack} To further demonstrate the consistent robustness of \ours across modalities, we conduct a single-modality attack experiment on CMU-MOSEI using the PGD attack. As shown in Figure~\ref{fig:balance}, \ours significantly improves robustness for high-vulnerability modalities while maintaining performance in others. 
This demonstrates that \ours effectively strengthens the defense of high-vulnerability modalities
, enhancing overall robustness.

\subsubsection{Efficiency of \ours} 
The most critical advantage of fast adversarial training lies in the efficiency of its single-step attack. To assess the computational overhead introduced by \ours, we measure the average training time per batch across different methods on CMU-MOSEI, as shown in Figure~\ref{fig:computational_time}.  The results show that the additional overhead remains constant and negligible across various methods. Meanwhile, as attack method becomes more complex or adopts multi-step attacks, the relative time cost introduced by \ours further decreases. This demonstrates that \ours improves robustness while maintaining high training efficiency.


\subsection{Ablation Study}

We conduct an ablation study on CMU-MOSEI dataset with different training strategies to demonstrate the effectiveness of \ours:
\begin{itemize}
    \item S1: Regularize only the most vulnerable modality in the current epoch, rather than applying regularization to all modalities simultaneously.
    \item S2: Utilize our vulnerability-aware attack to directly generate adversarial examples, where the perturbation budget for each modality may change in every epoch.
    \item S3: Attack only the most vulnerable modality, which is identified by applying the attack on the undefended model (i.e., Video modality in Figure~\ref{fig:attack_robustness}).
    \item Trap: Optimize the $\|\mathrm{x}\|_F \cdot \|\nabla_{\mathrm{x}}\mathcal{L}\|_F$, which may lead to overfitting and degraded expressive capability.
\end{itemize}

\begin{table}[t]
\footnotesize
\centering
\begin{tabular}{c @{\hskip 3pt} | @{\hskip 3pt} c @{\hskip 3pt} c @{\hskip 3pt} c @{\hskip 3pt} c}
\toprule
{Strategy} & FGSM-RS & FGSM-EP & FGSM-MEP & FGSM-PCO \\
\midrule
S1       & 62.63   & 66.79   & 65.15   & 63.45 \\
S2       & 35.54   & 17.51   & 52.14   & 26.60 \\
S3       & 51.88   & 59.96   & 62.76   & 64.98 \\
Trap     & 75.83   & 74.52   & 74.31   & 72.80 \\
\ours    & 68.49   & 68.58   & 67.46   & 65.58 \\
\bottomrule
\end{tabular}
\caption{Ablation of different adversarial training strategies under PGD attack ($\lambda = 0.5$).}
\label{tab:ablation_strategy}
\end{table}

\begin{table}[t]
\footnotesize
\centering
\begin{tabular}{c @{\hskip 3pt} | @{\hskip 3pt} c @{\hskip 3pt} c @{\hskip 3pt} c @{\hskip 3pt} c}
\toprule
Strategy & FGSM-RS & FGSM-EP & FGSM-MEP & FGSM-PCO \\
\midrule
Trap     & 7.15    & 29.51   & 13.74    & 2.76 \\
\ours    & 9.58    & 50.03   & 39.33    & 53.67 \\
\bottomrule
\end{tabular}
\caption{Robustness (\%) of Trap and \ours under the input-space PGD attack ($\lambda=0.1$).}
\label{tab:ablation_input}
\end{table}

As shown in Table~\ref{tab:ablation_strategy}, except for the Trap strategy, \ours consistently outperforms other baselines across different methods. This suggests that the proposed regularization effectively enhances overall robustness.
However, the Trap strategy maintains robustness under attack that is nearly equivalent to its performance on clean inputs, suggesting possible overfitting to the specific attack constraints.

To further investigate this phenomenon, we conduct an additional experiment in the input space. Specifically, we directly transfer the attack constraints from the feature space to the input space. Although this inevitably generates adversarial examples that may exceed the natural bounds of individual modalities, we argue that the results still provide meaningful insights into the underlying optimization trap.
As shown in Table~\ref{tab:ablation_input}, the Trap strategy exhibits severe overfitting across various fast adversarial training methods. This suggests that although feature magnitude may correlate with vulnerability in a frozen model, directly optimizing it leads to undesired overfitting and should be avoided.


\section{Conclusion}
In this paper, we reveal that inherent differences in modality-specific vulnerabilities have a significant impact on model robustness under adversarial attacks. We demonstrate that leveraging these differences enables the construction of more targeted and effective adversarial attacks. To address this issue, we propose \ours, a novel vulnerability-aware robust multimodal adversarial training method that introduces a regularization term to reduce and balance vulnerabilities across modalities. Experimental results on various multimodal datasets validate the effectiveness and efficiency of \ours, and further highlight the importance of differentiated training based on the unequal contributions of each modality to improving adversarial robustness.


\bibliography{aaai2026}

\end{document}